\documentclass[final,3p,times,twocolumn]{elsarticle}

\usepackage{amssymb}
\usepackage{amsmath}
\usepackage[numbers]{natbib}
\usepackage{algorithm}
\usepackage{algpseudocode} 
\usepackage{amsmath}
\usepackage{amssymb}
\usepackage{bm}
\usepackage{amsfonts}  
\usepackage{array}
\usepackage{booktabs}    
\usepackage{multirow}    
\usepackage{multicol}    
\usepackage{graphicx}      
\usepackage{booktabs}      
\usepackage{multirow}      
\usepackage{makecell}      
\usepackage{pifont}        
\usepackage{adjustbox}     
\usepackage{hyperref}
\hypersetup{
    colorlinks=true,      
    linkcolor=blue,       
    citecolor=red,        
    urlcolor=cyan         
}

\journal{Nuclear Physics B}

\begin{document}

\begin{frontmatter}



\title{FPC-VLA: A Vision-Language-Action Framework with a Supervisor for Failure Prediction and Correction}

\author{
    Yifan Yang$^{1,2}$, 
    Zhixiang Duan$^{2\dagger}$, 
    Tianshi Xie$^{2}$, 
    Fuyu Cao$^{3,2}$, 
    Pinxi Shen$^{2}$, 
    Peili Song$^{1}$, 
    Chenyang Zhao$^{1}$,\\
    Piaopiao Jin$^{2}$, 
    Guokang Sun$^{2}$, 
    Shaoqing Xu$^{2,4}$, 
    Yangwei You$^{2}$, 
    and Jingtai Liu$^{1*}$
}

\affiliation{This work is supported by the National Natural Science Foundation of China under Grant 62173189.}
\affiliation{This work is done during internship of Yifan Yang, Tianshi Xie, Fuyu Cao, Pinxi Shen at Xiaomi EV.}
\affiliation[1]{
    The Institute of Robotics and Automatic Information System, Tianjin Key Laboratory of Intelligent Robotics, and TBI Center, Nankai University, Tianjin 300350, China. 
    Emails: yangyifan@mail.nankai.edu.cn; liujt@nankai.edu.cn
}

\affiliation[2]{
    Xiaomi EV, Beijing, China
}

\affiliation[3]{
    Faculty of Robot Science and Engineering, Northeastern University, Shenyang 110819, China
}

\affiliation[4]{
    The State Key Laboratory of Internet of Things for Smart City, and Centre for Artificial Intelligence and Robotics, and Department of Electromechanical Engineering, University of Macau, Macau SAR, China
}
\affiliation{*Corresponding author. $\dagger$ Project leader.}

\begin{abstract}
Robotic manipulation is a fundamental component of automation. However, traditional perception-planning pipelines often fall short in open-ended tasks due to limited flexibility, while the architecture of a single end-to-end Vision-Language-Action (VLA) offers promising capabilities but lacks crucial mechanisms for anticipating and recovering from failure. To address these challenges, we propose FPC-VLA, a dual-model framework that integrates VLA with a supervisor for failure prediction and correction. The supervisor is triggered at keyframes and evaluates action viability through structured vision–language queries. If a potential failure is detected, it generates natural language corrections specifying direction and magnitude of adjustment. Additionally, we introduce an automated pipeline to generate large-scale failure prediction and correction datasets from existing robotic data without manual labeling. FPC-VLA also includes a dual-stream action fusion module that refines outputs by aggregating historical action predictions, using cosine similarity and temporal decay to weight past pose and gripper states separately. Evaluation results on multiple simulation platforms (SIMPLER and LIBERO) and robot embodiments (WidowX, Google Robot, Franka) show that FPC-VLA outperforms state-of-the-art models. Successful real-world deployments on Xiaomi Robot and ALOHA confirm FPC-VLA's strong generalization and practical utility for building more reliable autonomous systems. Our project page is \href{https://fpcvla.github.io/}{https://fpcvla.github.io/}.
\end{abstract}

\begin{graphicalabstract}
\includegraphics[width=0.7\textheight, viewport=1499 1403 2267 2224, clip=true]{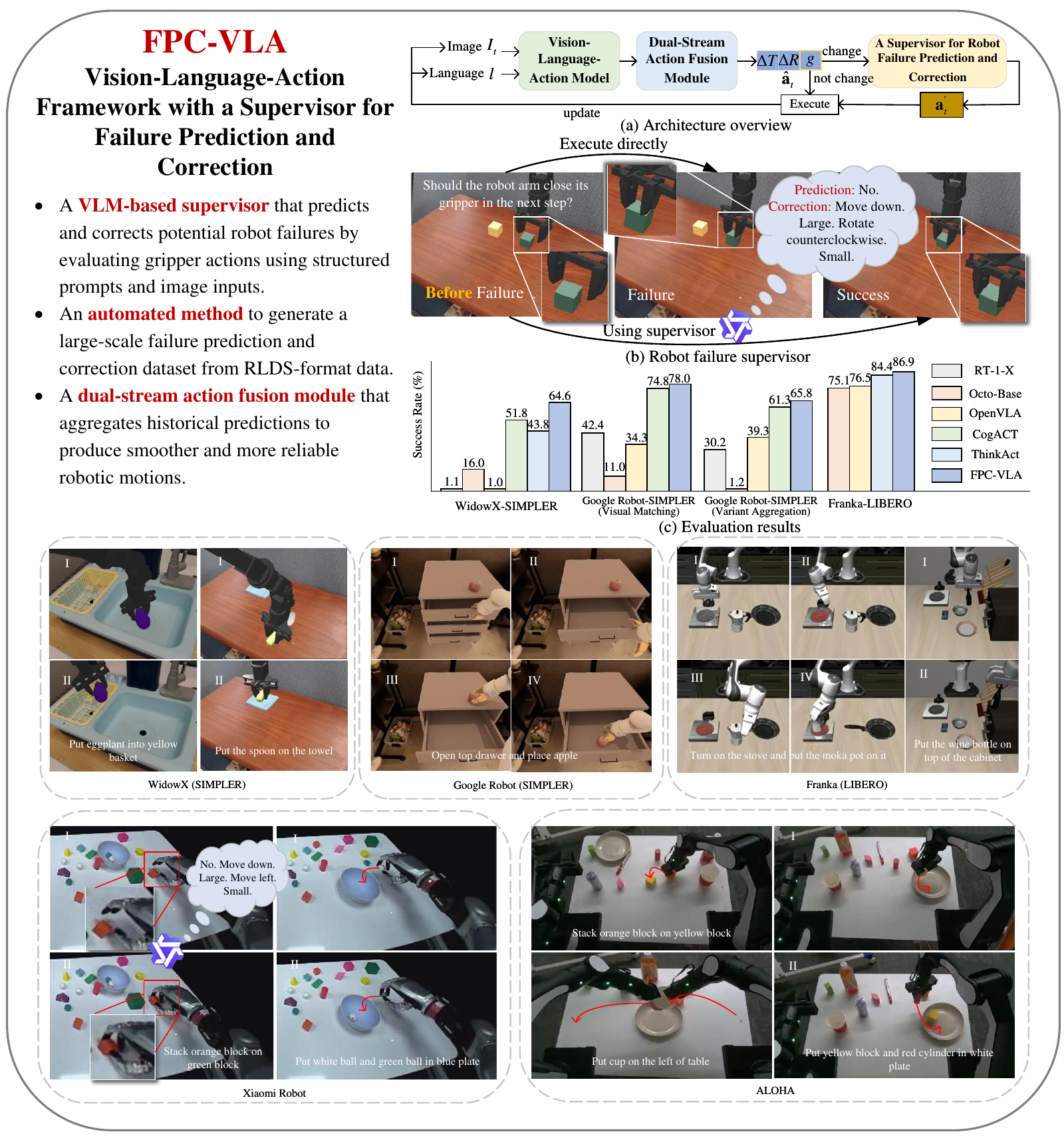}
\end{graphicalabstract}

\clearpage
\onecolumn
\begin{highlights}
\item A VLM-based supervisor that predicts and corrects potential robot failures by evaluating gripper actions using structured prompts and image inputs.
\item An automated method to generate a large-scale failure prediction and correction dataset from RLDS-format data.
\item A dual-stream action fusion module that aggregates historical predictions to produce smoother and more reliable robotic motions.
\item Evaluation results on SIMPLER and LIBERO benchmarks, using WidowX, Google Robot and Franka, and real-world experiments on Xiaomi robot and ALOHA, show that FPC-VLA outperforms other state-of-the-art methods. 
\end{highlights}

\clearpage
\twocolumn

\end{frontmatter}




\section{Introduction}

Robot manipulation serves as a cornerstone for advancing automation across a diverse range of sectors, including manufacturing, logistics, and healthcare \cite{10883018,10569055,LI2024111947}. In these domains, robots are increasingly expected to perform complex tasks such as assembly, sorting, and even delicate surgical operations, which demand high precision and reliability. Conventional robotic systems typically adopt a well-established “perception-planning” pipeline, which decomposes the manipulation process into two sequential stages. The first stage involves perception, where the system estimates the pose of target objects and identifies predefined grasp points using sensors and computer vision algorithms \cite{yang2024ps6d, yang2025mkpose,LIU2025130435,PAN2025132023,LI2023110491}. The second stage focuses on planning, where motion trajectories are generated to guide the robot to execute the grasp or manipulation action effectively \cite{DENG2025130204,HUANG2025113738}.

\begin{figure}[tb]
	\centering
    \includegraphics[width=1\linewidth, viewport=987 1277 1432 1643, clip=true]{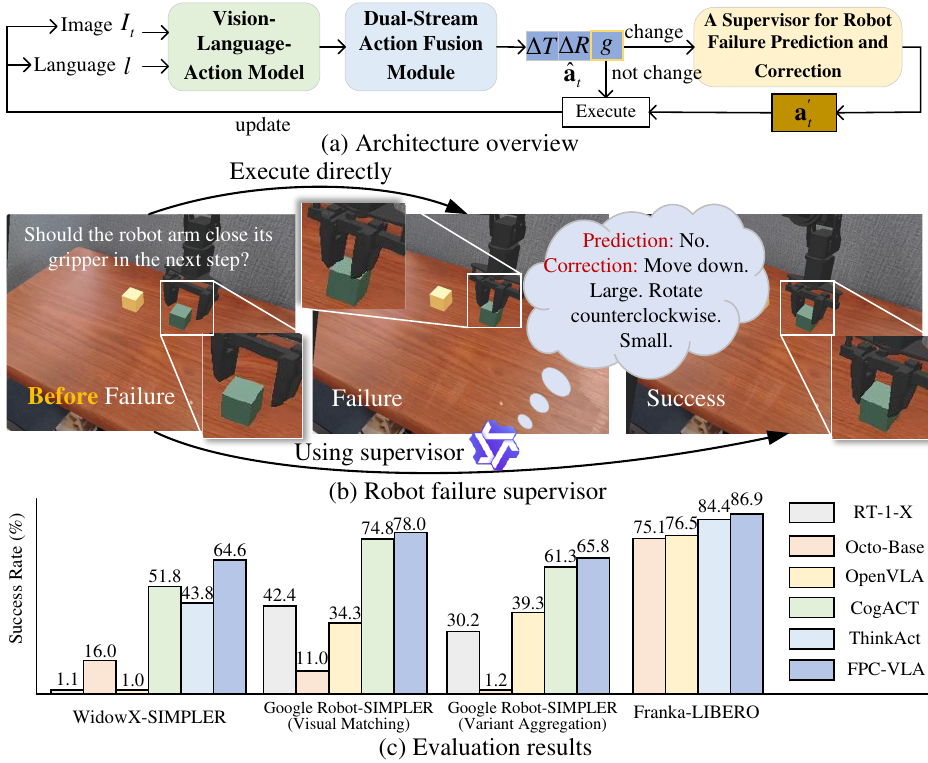}
	\caption{We propose FPC-VLA, a dual-model framework that integrates VLA with a supervisor for failure prediction and correction. It outperforms other advanced methods in all benchmarks. 
}
    \label{challenge}
\end{figure}

While this two-stage framework has proven effective in structured environments, it relies heavily on explicit object recognition and accurate geometric modeling, often assuming full prior knowledge of the target objects \cite{yang2024ps6d, yang2025mkpose}. This inherent dependency significantly restricts the system's flexibility and adaptability, particularly in open-ended or dynamically changing tasks where objects may be unknown or non-standard. Moreover, the performance of the perception module is frequently compromised under challenging conditions such as partial occlusion, cluttered scenes, or when encountering novel object shapes. In such cases, both pose estimation and grasp-point detection are prone to failures, leading to erroneous perceptual outputs. These inaccuracies inevitably propagate to the subsequent planning phase, resulting in unreliable or even failed robotic actions. Therefore, overcoming these limitations is critical for developing more robust and generalizable robot manipulation systems capable of operating in real-world, unstructured settings.

Unlike traditional modular pipelines that decompose robotic control into sequential components, Vision–Language–Action (VLA) models \cite{brohan2022rt,zitkovich2023rt,o2024open,team2024octo,kim2024openvla,qu2025spatialvla,zheng2024tracevla,li2024towards,li2024cogact, bjorck2025gr00t, zhang2024grape} integrate perception, task reasoning, and motion synthesis into a single, end-to-end trainable architecture. These models map raw visual observations and natural language instructions directly into low-level robot actions, eliminating the need for intermediate representations such as object detections or predefined state machines. Leveraging large-scale pretraining on diverse vision-language datasets, VLAs acquire a robust understanding of scenes and instructions, enabling strong generalization across a variety of objects. Building on advances in Vision-Language Models (VLM) \cite{CHOI2025113986,ZHENG2026129846,ZHANG2025128577,ZHAO2024124926} and  Vision-Language Navigation (VLN) \cite{WEN2024112610,WU2025112735,WU2025126442}, VLA models exhibit enhanced spatial and contextual reasoning, allowing them to dynamically infer task intent from language and continuously align their actions with human instructions. 

Despite the significant advances in VLAs, there are still some limitations. First, as illustrated in Fig. \ref{challenge}(b), the majority of existing approaches \cite{brohan2022rt,zitkovich2023rt,o2024open,team2024octo,kim2024openvla,qu2025spatialvla,zheng2024tracevla,li2024towards,li2024cogact, bjorck2025gr00t, zhang2024grape} predominantly rely on imitation learning from expert demonstrations, which are typically composed solely of successful execution trajectories. This leads to a fundamental weakness: these models struggle to recover from deviations, often requiring external supervision for potential failure prediction and corrective guidance. Moreover, collecting large-scale datasets that capture both failure prediction and correction is another bottleneck. Most existing datasets usually annotate failures post hoc \cite{duan2024aha, luo2025roboreflect, dai2024racer, lu2025robofac}, and rarely enable early prediction or proactive interventions. The lack of a standardized or scalable methodology limits the model’s generalization ability. In addition, many current methods segment actions into action chunks but execute only a portion of it \cite{brohan2022rt,o2024open,team2024octo,kim2024openvla}, losing the information of the latter part of the predicted sequence. Simply averaging actions from different modes, likely resulting in an action that does not belong to any of the correct action modes, overlooks the multimodal nature of action trajectories—i.e., the fact that there may be multiple plausible ways to achieve a goal. 

To address these challenges, we propose FPC-VLA, a vision-language-action framework with a supervisor for failure prediction and correction, as illustrated in Fig. \ref{challenge}(a). Unlike conventional end-to-end VLA models that directly execute predicted actions without verification, FPC-VLA introduces a VLM-based supervisor that actively evaluates the feasibility and safety of proposed actions before execution. If a potential failure is anticipated, the supervisor generates structured, directional corrections, which are then used to refine the original actions. We also introduce an automated method to generate a large-scale failure prediction and correction dataset from RLDS-format data, pairing images, task instructions, and gripper-focused QA prompts. Furthermore, to enhance the robustness and temporal consistency of the primary VLA's actions, we propose a dual-stream action fusion module. This module aggregates a history of past action predictions, intelligently weighting them based on their cosine similarity to the current prediction and applying a temporal decay to prioritize recent inputs. Crucially, this fusion is performed with decoupled pose and gripper state streams, acknowledging their distinct behavioral semantics. The result is a smoother, more reliable action sequence that is less susceptible to erratic outputs and better leverages the multi-modal nature of valid action trajectories, surpassing other state-of-the-art algorithms as demonstrated in Fig. \ref{challenge}(c).

The main contributions of this work are as follows:

\begin{enumerate}
    \item We propose a VLM-based supervisor that predicts and corrects potential robot failures by evaluating gripper actions using structured prompts and image inputs.
    \item We introduce a dual-stream action fusion module that aggregates historical predictions to produce smoother and more reliable robotic motions.
    \item We evaluate FPC-VLA on SIMPLER and LIBERO benchmarks, using WidowX, Google Robot and Franka, and test it for real-world applicability on Xiaomi robot and ALOHA. The results show that FPC-VLA outperforms other state-of-the-art methods.
\end{enumerate}

\section{Related Work}
\subsection{Traditional Perception-Planning Pipelines for Robotic Grasping}
Conventional robotic manipulation systems typically adopt a modular, two-stage pipeline that decouples perception and action planning. In the perception module, many approaches perform 6-degree-of-freedom (6-DoF) pose estimation of target objects to ensure that the gripper can execute subsequent grasping tasks with the correct orientation. In robot bin-picking tasks, \cite{yang2024ps6d} proposes an instance-level pose estimation method that, given known object CAD models, accomplishes centroid regression and rotation prediction for randomly stacked workpieces, effectively handling both highly symmetric and slender objects. UPG \cite{LI2023110491} leverages U-disparity continuity for rapid scene partitioning and employs an enhanced PointNet++ architecture to identify the topmost object in cluttered piles. Building upon \cite{yang2024ps6d}, \cite{yang2025mkpose} introduces a multimodal category-level pose estimation approach that takes RGB-D data and category-level textual descriptions as input, addressing pose estimation for known categories without requiring CAD models. FS-Gen6D \cite{PAN2025132023} enhances object localization under sparse observations through a multi-head dynamic detection mechanism and achieves robust, generalizable pose prediction by integrating a cross-stage residual architecture with deformable 3D convolutions. \cite{DENG2025130204} introduces an 8-DoF robotic hand platform and demonstrates that combining reward shaping with domain randomization in PyBullet enables rapid simulation-to-real transfer of reinforcement learning policies. \cite{LIU2025130435} introduces a multi-modal framework for human–robot handover that fuses speech, syntax, and vision for intent recognition, uses point clouds for 6-DoF grasp planning, and employs an improved DMP for human-like trajectory generation. 
\subsection{Vision-Language Models}
Recent advances in Vision-Language Models (VLMs) have significantly enhanced the perceptual and reasoning capabilities, offering powerful tools to bridge visual observations with linguistic instructions. A prominent line of work focuses on prompt learning to better align vision and language representations for specific tasks. For instance, \cite{ZHENG2026129846} introduces Group-guided Prompt Learning, a prompt learning framework that leverages LLMs to extract and incorporate class-level group knowledge into textual prompts. \cite{ZHANG2025128577} proposes a distillation-based prompt learning method that reuses soft supervision across training stages and introduces a region-aware dual prompt space with positive-negative mutual learning. Knowledge distillation has proven instrumental in transferring capabilities from large, general-purpose multimodal teachers to compact, task-efficient student models. \cite{CHOI2025113986} combines unimodal encoders with a lightweight fusion module, then boosts performance via knowledge distillation from a larger multimodal teacher. And also, \cite{ZHAO2024124926} proposes a few-shot object detection method that uses vision-language models to discover high-quality unlabeled novel instances in base-class data and treats ambiguous base images as implicit sources of novel-class semantics. These advances highlight VLMs’ growing role not just in perception, but in semantic reasoning and data-efficient learning capabilities, which provide a strong foundation for our use of a VLM-based supervisor for proactive failure prediction and correction.
\subsection{Vision-Language-Action Models}
Recent advances in multi-modal learning have driven the development of VLAs. Notable early examples include RT-1 \cite{brohan2022rt}, RT-2 \cite{zitkovich2023rt}, and RT-X \cite{o2024open}, which enable end-to-end training on large datasets, achieving strong generalization across platforms. Open-source models like Octo \cite{team2024octo} and OpenVLA \cite{kim2024openvla} have also emerged, pre-trained on extensive robot datasets and fine-tuned with domain-specific data. More recent efforts, such as SpatialVLA \cite{qu2025spatialvla} with Ego3D encoding and adaptive action grids, and TraceVLA \cite{zheng2024tracevla} with visual trace prompting, enhance spatial and spatiotemporal reasoning. Other developments include RoboVLMs \cite{li2024towards}, which use continuous outputs and policy heads, and CogACT \cite{li2024cogact}, which employs diffusion action transformers for modeling action sequences. Despite these advances, VLA models still struggle with autonomously detecting and recovering from failures in unstructured environments.

\begin{figure*}[tb]
  \centering
  \includegraphics[width=1\linewidth, viewport=25 1274 600 1580, clip=true]{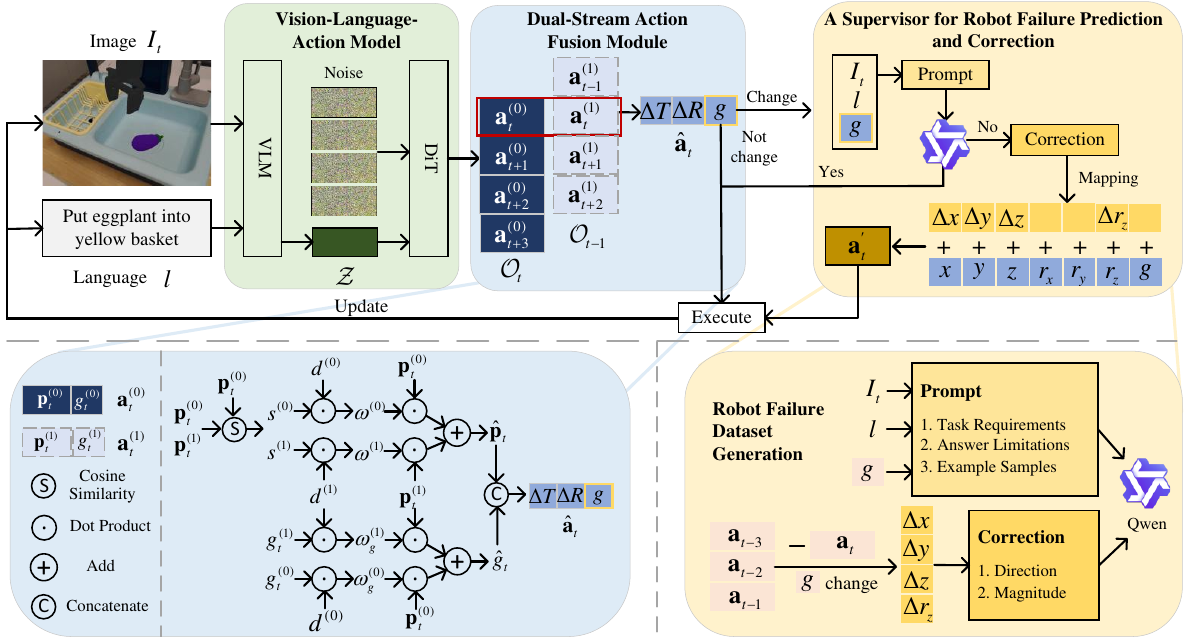}
  \caption{Architecture of FPC-VLA. 
The framework takes as input an observed image and a natural language instruction. A Vision-Language-Action (VLA) model first predicts a sequence of actions, which is then refined by a Dual-Stream Action Fusion Module that integrates historical predictions with the current prediction to generate the end-effector pose increment and the gripper state of the robotic arm. At keyframes where the gripper state changes, a VLM-based Supervisor detects potential failures and provides corrective guidance to optimize the action execution.}
  \label{network}
\end{figure*}

\subsection{Robot Failure Detection and Correction}
To address robot failure challenges, recent studies have explored using Multimodal Large Language Models (MLLMs) as auxiliary agents for error detection and reasoning. \cite{duan2024aha} can identify the reasons for failure, but does not provide a correction strategy. \cite{xiaoyang2024yell} implements failure correction but relies on human-provided instructions. \cite{luo2025roboreflect, dai2024racer, lu2025robofac} are able to autonomously adjust strategies after grasping failures, but lack the ability to anticipate such failures beforehand. Among them, RACER \cite{dai2024racer} also uses an external supervisor, but it only provides a corrected task instruction in natural language after a failure, which cannot be directly mapped to action changes. EvolvingGrasp \cite{zhu2025evolvinggrasp} improves through preference alignment from both successes and failures, and Phoenix \cite{xia2025phoenix} connects semantic reflection with motion-level correction. Unlike the approaches primarily focusing on post-failure adaptations or manual intervention, our approach anticipates failure scenarios and provides preemptive corrective actions.

\section{Method}
\subsection{Architecture Overview and Problem Formulation}
As shown in Fig. \ref{network}, the input to FPC-VLA consists of an RGB image $I_t \in \mathbb{R}^{H \times W \times 3}$ and a language instruction $l$ at timestep $t$ . The output is a refined action $\mathbf{a}^{'}_t \in \mathbb{R}^7$ that corrects failure-prone predictions. The proposed pipeline includes three modules: 1) a robot failure dataset generation method, 2) a VLM-based supervisor for failure prediction and correction, and 3) a dual-stream action fusion module.

In the dataset generation module, grasp events are identified by detecting changes in gripper state $g_t$, forming a change set $\mathcal{C}$. For each event $c \in \mathcal{C}$, we retain frames $\mathcal{R}_c $, and compute pose differences $\Delta \mathbf{a}$ . These are converted into structured language descriptions using thresholds $\delta_{\text{small}}, \delta_{\text{large}}$, and combined with task and answer constraints to form prompts $\mathcal{P}_t$. In the dual-stream action fusion module, a set of past predictions $\mathcal{A}_t$ is collected, each decomposed as $\mathbf{a}_t^{(k)} = [\mathbf{p}_t^{(k)}, g_t^{(k)}]$ in observation $\mathcal{O}_{t-k}$. Cosine similarity $\text{sim}^{(k)}$ with the latest pose $\mathbf{p}_t^{(0)}$ is smoothed via sigmoid $s^{(k)}$, weighted by time decay $d^{(k)}$, and normalized to produce fusion weights $w^{(k)}$, yielding the fused action $\hat{\mathbf{a}}_t = [\hat{\mathbf{p}}_t, \hat{g}_t]$. In the VLM-based failure prediction and correction module, when $g_t$ changes, the model $\mathcal{M}$ takes image $I_t$ and prompt $\mathcal{P}_t$ as input, returning response $R_t = \mathcal{M}(I_t, \mathcal{P}_t)$. If $R_t$ starts with “No”, translation correction $\Delta \mathbf{p} \in \mathbb{R}^3$ and rotation correction $\Delta r_z \in \mathbb{R}$ are extracted to obtain refined action $\mathbf{a}'_t$.

\begin{figure*}[tb]
  \centering
  \includegraphics[width=0.9\linewidth, viewport=62 1763 614 1935, clip=true]{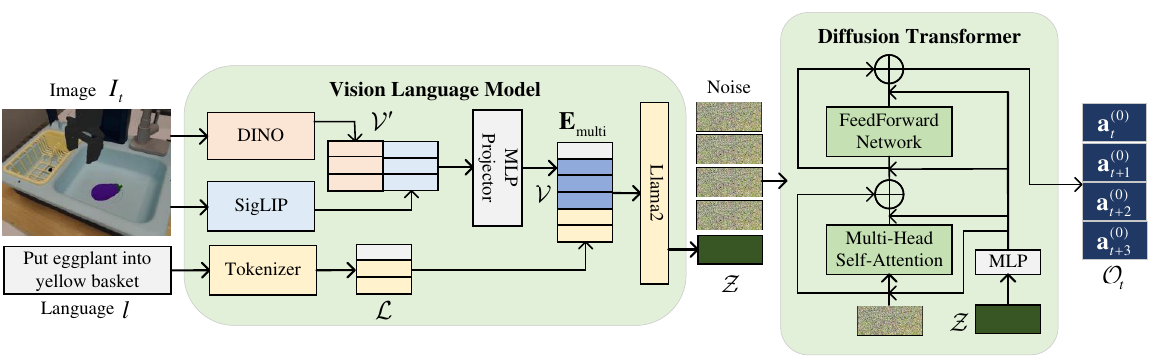}
  \caption{Architecture of Vision-Language-Action Model. The input image passes through two visual feature encoders and an MLP projector, is concatenated with tokenized language features, and then fed into Llama2 to obtain cognitive features. These features serve as conditioning inputs to a Diffusion Transformer, which progressively denoises noise to generate a predicted action sequence conditioned on the current observation. }
  \label{vla}
\end{figure*}

\subsection{Vision-Language-Action Model}
\noindent\textbf{Vision Module \quad}The vision module is responsible for extracting patch-level features from input images and projecting them into the embedding space of the language model. As shown in Fig. \ref{vla}, the vision module employs a dual-encoder architecture that processes images through both DINO \cite{oquab2023dinov2} and SigLIP \cite{oquab2023dinov2} vision backbones, followed by feature fusion. Given image $I_t$, it is passed through hybrid visual encoder and then passed through a linear projection to align with the LLM embedding space to get vision tokens $\mathcal{V} = \{v_1, v_2, ..., v_{N_\mathcal{V}}\}, \  \mathcal{V} \in \mathbb{R}^{N_{\mathcal{V}} \times d}$, 
\begin{equation}
    \mathcal{V} = \mathbf{W}_p \ \text{Concat}(f_{\text{dino}}(I_t), f_{\text{siglip}}(I_t)) + \mathbf{b}_p,
\end{equation}
where \( \mathbf{W}_p \in \mathbb{R}^{d \times (d_{\text{dino}} + d_{\text{siglip}})} \) is the projection matrix and \( d \) is the language model's hidden dimension. $\mathbf{b}_p \in \mathbb{R}^d$ is the bias vector of the projection layer.

\noindent\textbf{Language Module \quad}The language module is built on Llama2 \cite{touvron2023llama}, a transformer-based architecture designed for autoregressive language modeling. The input \( l \) is tokenized through Llama tokenizer and get a sequence of language tokens $\mathcal{L} = \{l_1, l_2, \ldots, l_{N_\mathcal{L}}\}$, where each token $l_i$ is mapped to a high-dimensional embedding space $\mathbb{R}^d$. For multimodal inputs, the text embeddings $\mathcal{L}$ are combined with visual embeddings $\mathcal{V}$ from the vision module
\begin{equation}
    \mathbf{E}_{\text{multi}} = \text{Concat}\left(\mathcal{L}_{\text{BOS}}, \mathcal{V}, \mathcal{L}\right),
\end{equation}
where BOS token is the beginning-of-sequence token. The output cognition token $\mathcal{Z} \in \mathbb{R}^d$ is used as input for the action module to interpret and generate the required actions.

\noindent\textbf{Action Module \quad}The action module uses Diffusion Transformer (DiT) \cite{peebles2023scalable} to generate action sequences conditioned on $\mathcal{Z}$. Diffusion model generates actions by gradually denoising a noisy input, while the transformer backbone enhances the model’s ability to capture long-range dependencies in the data. 

The forward diffusion process is a fixed Markov chain that gradually adds noise to the ground-truth action sequence $x_0$ over $T$ timesteps. During training, a noisy sample $x_t$ at an arbitrary timestep $t$ is efficiently obtained via the $q$-sampling process
\begin{equation}
x_t = \sqrt{\bar{\alpha}_t} \cdot x_0 + \sqrt{1 - \bar{\alpha}_t} \cdot \epsilon, \ \epsilon \sim \mathcal{N}(\mathbf{0}, \mathbf{I}),
\end{equation}
where $\bar{\alpha}_t$ is the cumulative product of the noise schedule $\alpha_t$ and controls the noise level at timestep $t$, and $x_0$ represents the ground-truth action sequence for the subsequent $n$ steps. The DiT model is trained to learn the reverse denoising process. Specifically, given the noisy input $x_t$, the timestep $t$, and the conditioning variable $\mathcal{Z}$, the model is tasked to predict the noise component $\epsilon$ that was added to $x_0$
\begin{equation}
\text{DiT}(x_t, t, \mathcal{Z}) \approx \epsilon.
\end{equation}
The architecture of the DiT block is pivotal to its performance. The input action sequence $x_t$ is first patchified and projected into a sequence of embeddings. The timestep $t$ and the conditioning variable $\mathcal{Z}$ are incorporated into the network through Adaptive Layer Norm (AdaLN), where the parameters for layer normalization are dynamically generated from the embeddings of $t$ and $\mathcal{Z}$. The core of the block consists of Multi-Head Self-Attention and Feed-Forward Network layers, which enable the model to effectively capture the global dependencies that are crucial for coherent multi-step action generation.

During inference, the process starts from a random noise vector $x_T \sim \mathcal{N}(\mathbf{0}, \mathbf{I})$. The trained DiT model then iteratively denoises this vector over $T$ steps, conditioned on $\mathcal{Z}$, to produce a plausible and context-aware action sequence $x_0$.

\subsection{Dual-Stream Action Fusion Module} \label{E}
To obtain a robust and smooth action output and avoid waste in action prediction, we employ a dual-stream action fusion strategy. At each timestep $t$, the output action $\hat{\mathbf{a}}_t$ is not solely determined by the most recent policy output, but rather by fusing a set of historical predictions, each made from earlier observations. This strategy fully takes into account the multimodality and temporal properties of actions, as well as the differences in pose and gripper states.

Let $\mathcal{O}_{t-k}$ denote the observation at time $t-k$, and let $\mathbf{a}_t \mid \mathcal{O}_{t-k}$ represent the model's prediction for the action at time $t$, based solely on the observation available at time $t-k$. We denote this prediction as
\begin{equation}
\mathbf{a}_t^{(k)} := \mathbf{a}_t \mid \mathcal{O}_{t-k}.
\end{equation}
We collect a set of such predictions from the past $N$ time steps
\begin{equation}
\mathcal{A}_t = \left\{ \mathbf{a}_t^{(k)} \mid k = 0, 1, \dots, N-1 \right\}.
\end{equation}

\begin{algorithm}[H]
\small
\caption{Dual-Stream Action Fusion Strategy}
\label{alg:action_fusion}
\begin{algorithmic}
\Require 
    $\text{VLA}(\cdot)$ , $\{\mathcal{O}_{t-k} \mid k = 0, 1, \dots, N-1\}$ , $\lambda > 0$, $\alpha > 0$, $\beta \in [0, 1)$

\Ensure Fused action: $\hat{\mathbf{a}}_t = [\hat{\mathbf{p}}_t, \hat{g}_t] \in \mathbb{R}^7$

\State Initialize $\mathcal{A}_t \leftarrow \emptyset$, $\mathcal{P}_t \leftarrow \emptyset$, $\mathcal{G}_t \leftarrow \emptyset$

\For{$k = 0$ to $N-1$}
    \State Predict action using $\mathcal{O}_{t-k}$: $\mathbf{a}_t^{(k)} \leftarrow \text{VLA}(\mathcal{O}_{t-k})$
    \State Decompose action: $[\mathbf{p}_t^{(k)}, g_t^{(k)}] \leftarrow \mathbf{a}_t^{(k)}$
    \State $\mathcal{A}_t \leftarrow \mathcal{A}_t \cup \{\mathbf{a}_t^{(k)}\}$, $\mathcal{P}_t \leftarrow \mathcal{P}_t \cup \{\mathbf{p}_t^{(k)}\}$, $\mathcal{G}_t \leftarrow \mathcal{G}_t \cup \{g_t^{(k)}\}$
\EndFor

\For{$k = 0$ to $N-1$}
    \State $\text{sim}^{(k)} \leftarrow \dfrac{\mathbf{p}_t^{(k)} \cdot \mathbf{p}_t^{(0)}}{\|\mathbf{p}_t^{(k)}\| \cdot \|\mathbf{p}_t^{(0)}\| + \varepsilon}$ \Comment{Cosine Similarity}
    \State Apply sigmoid transformation: $s^{(k)} \leftarrow \sigma(\alpha \cdot \text{sim}^{(k)})$
\EndFor

\For{$k = 0$ to $N-1$}
    \State $d^{(k)} \leftarrow \exp(-\lambda \cdot k)$ \Comment{Time Decay}
\EndFor

\State $S \leftarrow 0$
\For{$k = 0$ to $N-1$}
    \State $S \leftarrow S + \max(0, (1 - \beta) \cdot s^{(k)} \cdot d^{(k)})$ \Comment{Regularization}
\EndFor
\For{$k = 0$ to $N-1$}
    \State $w^{(k)} \leftarrow \dfrac{\max(0, (1 - \beta) \cdot s^{(k)} \cdot d^{(k)})}{S}$
\EndFor

\State $\hat{\mathbf{p}}_t \leftarrow \mathbf{0} \in \mathbb{R}^6$
\For{$k = 0$ to $N-1$}
    \State $\hat{\mathbf{p}}_t \leftarrow \hat{\mathbf{p}}_t + w^{(k)} \cdot \mathbf{p}_t^{(k)}$ \Comment{Pose Fusion}
\EndFor

\State $D \leftarrow \sum_{j=0}^{N-1} d^{(j)}$
\For{$k = 0$ to $N-1$}
    \State $\tilde{d}^{(k)} \leftarrow \dfrac{d^{(k)}}{D}$
\EndFor

\State $\hat{g}_t \leftarrow 0$
\For{$k = 0$ to $N-1$}
    \State $\hat{g}_t \leftarrow \hat{g}_t + \tilde{d}^{(k)} \cdot g_t^{(k)}$ \Comment{Gripper State Fusion}
\EndFor

\State $\hat{\mathbf{a}}_t \leftarrow [\hat{\mathbf{p}}_t, \hat{g}_t]$

\State \Return $\hat{\mathbf{a}}_t$
\end{algorithmic}
\end{algorithm}

Each $\mathbf{a}_t^{(k)} \in \mathbb{R}^7$ is decomposed as $\mathbf{a}_t^{(k)} = \left[ \mathbf{p}_t^{(k)}, g_t^{(k)} \right]$, where $\mathbf{p}_t^{(k)} \in \mathbb{R}^6$ represents the relative end-effector pose, and $g_t^{(k)} \in \mathbb{R}$ denotes the scalar gripper state. 

Using the most recent prediction $\mathbf{p}_t^{(0)}$ as a reference, we calculate the cosine similarity between each $\mathbf{p}_t^{(k)}$ and $\mathbf{p}_t^{(0)}$
\begin{equation}
\text{sim}^{(k)} = \frac{ \mathbf{p}_t^{(k)} \cdot \mathbf{p}_t^{(0)} }{ \| \mathbf{p}_t^{(k)} \| \cdot \| \mathbf{p}_t^{(0)} \| + \varepsilon },
\end{equation}
where $\varepsilon = 10^{-7}$ is a small constant for numerical stability. To smooth the similarity response and ensure differentiability, we apply a sigmoid transformation with a scaling factor $\alpha > 0$
\begin{equation}
s^{(k)} = \sigma(\alpha \cdot \text{sim}^{(k)}).
\end{equation}

This yields similarity weights $s^{(k)} \in (0,1)$ that reflect the semantic alignment between the current and past predictions. We intentionally exclude $g_t$ from the cosine similarity computation due to semantic misalignment, magnitude disparity, and behavioral decoupling — as the gripper often functions independently of the end-effector pose (e.g., holding an object steady while the arm moves).

To give higher importance to recent predictions, we apply an exponential decay over time
\begin{equation}
d^{(k)} = \exp(-\lambda \cdot k),
\end{equation}
where $\lambda > 0$ is a decay coefficient. The total weight $w^{(k)}$ for each pose is then computed as the product of similarity and temporal decay
\begin{equation}
w^{(k)} = \frac{\max(0,\ (1 - \beta) \cdot s^{(k)} \cdot d^{(k)})}{\sum_{j=0}^{N-1} \max(0,\ (1 - \beta) \cdot s^{(j)} \cdot d^{(j)})},
\end{equation}
where $\beta \in [0, 1)$ is a regularization factor that prevents overconfidence in a small subset of predictions.

The final fused pose  is computed as the weighted average
\begin{equation}
\hat{\mathbf{p}}_t = \sum_{k=0}^{N-1} w^{(k)} \cdot \mathbf{p}_t^{(k)}.
\end{equation}

For the scalar gripper state $g_t$, we intentionally do not apply similarity-based weighting, because the cosine similarity measure is computed based on $\mathbf{p}_t$, and does not meaningfully reflect the behavior or intent of $g_t$. Instead, we use temporal decay alone to perform a weighted average over historical gripper values
\begin{equation}
\hat{g}_t = \sum_{k=0}^{N-1} \tilde{d}^{(k)} \cdot g_t^{(k)}, \quad \text{where } \tilde{d}^{(k)} = \frac{d^{(k)}}{ \sum_{j=0}^{N-1} d^{(j)} }.
\end{equation}

The final output action $\hat{\mathbf{a}}_t$ is a concatenation of the fused pose and gripper state
$
\hat{\mathbf{a}}_t = [
\hat{\mathbf{p}}_t, 
\hat{g}_t
]
\in \mathbb{R}^7$.

\subsection{Robot Failure Prediction and Correction Dataset} \label{dataset}
Most existing large-scale VLA datasets \cite{o2024open, walke2023bridgedata, brohan2022rt, liu2023libero} only include successful trajectories, while current failure correction datasets \cite{luo2025roboreflect, lu2025robofac} are limited in scale and lack predictability. To address this, we propose an automated method for generating failure correction datasets from large-scale RLDS-format data and adapt them for VLM training. In practical experiments, in addition to synthetic data, simulation platforms such as MuJoCo and teleoperation methods with real robots are used to construct failure correction data through annotation.


Denote the dataset $\mathcal{D}$ as a collection of episodes. Each episode $\mathcal{E}_i$ consists of a sequence of steps
\begin{equation}
     \mathcal{D} = \{ \mathcal{E}_1, \mathcal{E}_2, \dots, \mathcal{E}_N \},\ 
     \mathcal{E}_i = \{ s_1, s_2, \dots, s_T \}.
\end{equation}
At each timestep $t$, the step $s_t$ encapsulates the full state of the system, defined by the tuple
\begin{equation}
s_t = (I_t, \mathbf{a}_t, l),
\end{equation}
where each step $s_t$ includes:
\begin{itemize}
    \item An RGB image observation $I_t \in \mathbb{R}^{H \times W \times 3}$,
    \item An action vector $\mathbf{a}_t \in \mathbb{R}^7$, typically comprising a 6-DOF end-effector pose change and a 1-D gripper state.
    \item A task-level language instruction $l$.
\end{itemize}

We posit that grasp and release events are pivotal moments in manipulation tasks, often signifying the transition between task phases. Let $g_t \in [0, 1]$ be the normalized gripper state at time $t$ (e.g., 0 for open, 1 for closed). The set of all gripper state change timesteps $\mathcal{C}$ is defined as
\begin{equation}
\mathcal{C} = \left\{ t \in \{2, \dots, T\} \ \middle| \ |g_t - g_{t-1}| > \delta_g \right\},
\end{equation}
where $\delta_g$ is a sensitivity threshold, empirically set to $\delta_g = 0.5$. To provide sufficient contextual information for a VLM to understand the scene dynamics leading up to a change, we retain a window of frames around each event. For a given change timestep $c \in \mathcal{C}$, the set of retained indices is
\[
\mathcal{R}_c = \{ c - k, \dots, c - 1, c \} \cap [1, T],
\]
where $k$ is the context window size. In our implementation, $k=3$. The union of all these sets forms the complete set of frames used for dataset generation
\[
\mathcal{R} = \bigcup_{c \in \mathcal{C}} \mathcal{R}_c.
\]

For every retained frame $s_t$ where $t \in \mathcal{R}$, we generate a corrective action target. This is achieved by identifying the next gripper change event that the robot must execute. Formally, we find:
\[
c^* = \min \{ c \in \mathcal{C} \mid c \ge t \}.
\]
The corrective action vector $\Delta \mathbf{a}$ is then computed as the difference between the action at this future event and the current action:
\[
\Delta \mathbf{a} = \mathbf{a}_{c^*} - \mathbf{a}_t.
\]
This vector $\Delta \mathbf{a}$ encodes the necessary pose adjustment to progress towards the next critical phase of the task. 

As indicated by the pre-training process of VLA \cite{zheng2024tracevla,li2024towards,li2024cogact,qu2025spatialvla}, directly obtaining continuous action corrections requires large-scale pre-training data and high training costs. Therefore, we map the continuous correction values into discrete natural language signals, thereby fulfilling the task requirements with only LoRA fine-tuning of the VLM. To make the corrective action comprehensible to a VLM, we map the continuous vector $\Delta \mathbf{a}$ into a structured natural language description. We decompose $\Delta \mathbf{a}$ into its translational ($\Delta x, \Delta y, \Delta z$) and rotational (primarily around the vertical axis, $\Delta r_z$) components.

Let $D_i$ be the direction for axis $i$ (e.g., `left', `right', `up', `down'), determined by the sign of $\Delta i$. The magnitude-based textual template for translation along axis $i$ is generated by the function:
\[
\text{Text}_{\text{trans}}(i) =
\begin{cases}
\text{None}, & |\Delta i| \le \delta_{\text{small}} \\
\text{Move $D_i$. Small}, & \delta_{\text{small}} < |\Delta i| \le \delta_{\text{large}} \\
\text{Move $D_i$. Large}, &  |\Delta i| > \delta_{\text{large}}
\end{cases}
\]

Similarly, the rotation around the $z$-axis is described by:
\[
\text{Text}_{\text{rot}}(z) =
\begin{cases}
\text{None}, &  |\Delta r_z| \le \delta_{\text{small}} \\
\text{Rotate $D_z$. Small}, & \delta_{\text{small}} < |\Delta r_z| \le \delta_{\text{large}} \\
\text{Rotate $D_z$. Large}, & |\Delta r_z| > \delta_{\text{large}}
\end{cases}
\]
where $D_z$ is either `clockwise' or `counter-clockwise'. Since rotations around the xy-axis are difficult to translate into natural language and VLMs have a poor understanding of them, here we only consider the most frequently occurring clockwise and counterclockwise rotation issues. In practice, $\delta_{\text{small}}=0.01$, $\delta_{\text{large}}=0.1$. The final instruction is a concatenation of all non-"None" components from $\text{Text}_{\text{trans}}(x)$, $\text{Text}_{\text{trans}}(y)$, $\text{Text}_{\text{trans}}(z)$, and $\text{Text}_{\text{rot}}(z)$. The specific mapping between variables and directional language is illustrated in Table \ref{Mapping}. Due to the consistent coordinate system in VLA training data such as \cite{walke2023bridgedata, brohan2022rt}, the prompts indicate that all directions are defined relative to the robot's base frame, rather than from the camera's perspective.

\begin{table}[tb]
	\renewcommand\arraystretch{0.8}
 \setlength{\abovecaptionskip}{0pt}    
        \setlength{\belowcaptionskip}{5pt}
        \setlength{\tabcolsep}{14pt}  
	\centering
	\caption{Mapping of variables and directions in BridgeV2 dataset}
	\resizebox{0.49\textwidth}{!}{
		\begin{tabular}{cccc}
\toprule
\textbf{Axis} & \textbf{Variable} & \textbf{Sign} & \textbf{Direction} $D$ \\
\midrule
\multirow{2}{*}{X}         & \multirow{2}{*}{$\Delta x$} & + & forward \\
                           &                         & $-$ & backward \\
\multirow{2}{*}{Y}         & \multirow{2}{*}{$\Delta y$} & + & left \\
                           &                         & $-$ & right \\
\multirow{2}{*}{Z}         & \multirow{2}{*}{$\Delta z$} & + & up \\
                           &                         & $-$ & down \\
\multirow{2}{*}{Around Z}  & \multirow{2}{*}{$\Delta r_z$}    & + & counterclockwise \\
                           &                         & $-$ & clockwise \\
\bottomrule
\end{tabular}}
	\label{Mapping}
\end{table}

Given the retained frame $s_t$, a standard prompt $\mathcal{P}_t$ consists of three parts: task requirements, answer limitations, and example samples. An example is shown in Table \ref{prompt}.
\begin{table}[tb]
    \setlength{\abovecaptionskip}{0pt}    
    \setlength{\belowcaptionskip}{5pt}
    \centering
    \caption{Example of failure prediction and correction prompts}
    \resizebox{0.49\textwidth}{!}{
        \begin{tabular}{p{2cm}p{6.5cm}}
            \toprule
            \textbf{Part} & \textbf{Content} \\
            \midrule
            Task Requirements & My task is to pick rxbar chocolate from bottom drawer and place on counter. Should the robot arm close its gripper in the next step? If not, please specify the direction and magnitude of the gripper's movement and rotation. \\
           
            Answer Limitations & The direction must be chosen from the following options: move up, move down, move left, move right, rotate clockwise, and rotate counterclockwise. At most one option can be selected from each pair: move up or move down, move left or move right, and rotate clockwise or rotate counterclockwise. The magnitude should be either large or small. Your response should start with `Yes' or `No'. If the answer is `No', then include the direction and magnitude of both movement and rotation. \\
            
            Example Samples & Here are some example responses: Example 1: Yes. Example 2: No. Move up. Large. Move left. Small. Example 3: No. Move down. Large. Rotate clockwise. Small. \\
            \bottomrule
        \end{tabular}
    }
    \label{prompt}
\end{table}
\subsection{A VLM-Based Supervisor for Robot Failure Prediction and Correction}
Existing robotic manipulation systems based on VLA models often directly execute predicted commands without assessing their contextual appropriateness. These systems typically lack the ability to predict the consequences of actions or to autonomously correct erroneous decisions. To address this limitation, we propose a supervisor for failure prediction and correction, which evaluates the suitability of gripper actions before execution. The algorithm workflow is shown in Algorithm \ref{alg:consistent_supervisor}.

\begin{algorithm}[tb]
\small
\caption{Supervisor for Failure Prediction and Correction}
\label{alg:consistent_supervisor}
\begin{algorithmic}[1]
\Require Vision-Language Model $\mathcal{M}$, Prompt Template $\mathcal{P}$, Threshold $\delta_g$
\State Initialize $g_{t-1} \gets 0$
\For{each timestep $t = 1, 2, \dots$}
    \State Receive $\hat{\mathbf{a}}_t = [\Delta x_t, \Delta y_t, \Delta z_t, \Delta r_{x,t}, \Delta r_{y,t}, \Delta r_{z,t}, g_t]^\top$ from action fusion module
    \State Receive current image observation $I_t$
    
    \If{$|g_t - g_{t-1}| > \delta_g$}
        \State $R_t \gets \mathcal{M}(I_t, \mathcal{P}_t)$
        
        \If{$R_t$ begins with "No"}
            \State Extract $\Delta \mathbf{p} = [\Delta x, \Delta y, \Delta z]$ and $\Delta r_z$ from $R_t$
            \State Map directional tokens to step sizes: $\delta_{\text{small}}$, $\delta_{\text{large}}$
            \State $\mathbf{a}'_t \gets \hat{\mathbf{a}}_t + [\Delta x, \Delta y, \Delta z, 0, 0, \Delta r_z, 0]$
            \State Execute $\mathbf{a}'_t$
        \Else
            \State Execute $\hat{\mathbf{a}}_t$
        \EndIf
    \Else
        \State Execute $\hat{\mathbf{a}}_t$
    \EndIf
    
    \State $g_{t-1} \gets g_t$
\EndFor
\end{algorithmic}
\end{algorithm}

Let the raw action command predicted by the primary VLA policy at timestep \( t \) be parameterized as a 7D vector:
\begin{equation}
\hat{\mathbf{a}}_t = [\Delta x_t, \Delta y_t, \Delta z_t, \Delta r_{x,t}, \Delta r_{y,t}, \Delta r_{z,t}, g_t]^\top,
\end{equation}
where \( (\Delta x, \Delta y, \Delta z) \in \mathbb{R}^3 \) denotes the desired end-effector positional displacement in the robot's base frame. \( (\Delta r_x, \Delta r_y, \Delta r_z) \in \mathbb{R}^3 \) represents the incremental rotation, and \( g_t \in \{0, 1\} \) is the binary gripper state command. This action vector \( \hat{\mathbf{a}}_t \) is the direct output from the action fusion module detailed in Section \ref{E}.

The framework checks gripper consistency by comparing the current $g_t$ and previous $g_{t-1}$. The robot failure prediction and correction supervisor is triggered when the difference exceeds the threshold $\delta_g$
\begin{equation}
     \left| g_t - g_{t-1} \right| > \delta_g.
     \label{g}
\end{equation}
A transition satisfying \eqref{g} signifies an intended change in the gripper's state, which is a high-stakes decision requiring contextual validation. When the gripper state $g_t$ meets the condition in \eqref{g}, the prompt $\mathcal{P}_t$ and the current image $I_t$ are input into a vision language model $\mathcal{M}$
\begin{equation}
R_t = \mathcal{M}(I_t, P_t),
\end{equation}
where $R_t$ represents the generated natural language response by the model $\mathcal{M}$.

\begin{figure*}[tb]
  \centering
  \includegraphics[width=1\linewidth, viewport=100 280 709 443, clip=true]{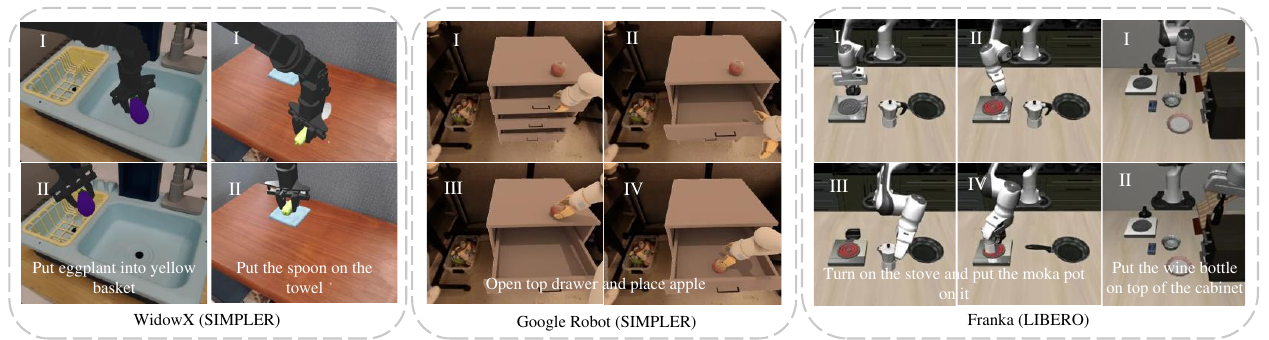}
  \caption{Simulation results of FPC-VLA on different robots. The model performs well even in long-horizon tasks and non-pick-and-place tasks (e.g., opening drawers and turning on the stove). }
  \label{viz}
\end{figure*}

If the response $R_t$ is ``Yes.", it indicates that $\mathcal{M}$ approves the decision made by VLA regarding the opening or closing of the gripper, and the original command \( \hat{\mathbf{a}}_t \) is executed without modification. If $R_t$ begins with ``No", the action $\hat{\mathbf{a}}_t$ needs to be adjusted according to the modification suggestions provided by $\mathcal{M}$. In the latter case, we extract structured directives from the text, including translation correction $\Delta \mathbf{p} \in \mathbb{R}^3$ and rotation correction $\Delta r_z \in \mathbb{R}$, based on predefined directional mappings. Each directional token is paired with a discrete magnitude indicator (small or large), which is mapped to a scalar step size.The refined action \( \mathbf{a}'_t \) is subsequently generated by applying the parsed correction to the original command, selectively overriding only the components for which a correction was specified
\begin{equation}
\mathbf{a}'_t = \hat{\mathbf{a}}_t + [\Delta x, \Delta y, \Delta z, 0, 0, \Delta r_z, 0].
\end{equation}
The refined action $\mathbf{a}'_t$ is then executed within the environment. 
\section{Experiment}

\begin{table*}[tb]
    \renewcommand\arraystretch{1}
    \setlength{\abovecaptionskip}{0pt}    
    \setlength{\belowcaptionskip}{5pt}
    \centering
    \caption{Comparison results on SIMPLER with WidowX}
    \begin{adjustbox}{width=\textwidth, center}
        \begin{tabular}{cccccccccccccc}
            \toprule
            \multirow{2}{*}{\textbf{Method}} & 
            \multirow{2}{*}{\makecell{\textbf{Fine-tuned}\\\textbf{(in-domain)}}} &
            \multicolumn{2}{c}{\makecell{\textbf{Put spoon}\\\textbf{on towel}}} & 
            \multicolumn{2}{c}{\makecell{\textbf{Put carrot}\\\textbf{on plate}}} & 
            \multicolumn{2}{c}{\makecell{\textbf{Stack green block}\\\textbf{on yellow block}}} & 
            \multicolumn{2}{c}{\makecell{\textbf{Put eggplant}\\\textbf{on yellow basket}}} & 
            \multirow{2}{*}{\textbf{Avg. Grasp}} & 
            \multirow{2}{*}{\textbf{Avg. Task}} \\
            & & Grasp & Task 
            & Grasp & Task 
            & Grasp & Task 
            & Grasp & Task 
            & & \\
            \midrule
            RT-1-X \cite{o2024open}& \ding{55} & 16.7 & 0.0 & 20.8 & 4.2 & 8.3 & 0.0 & 0.0 & 0.0 & 11.5 & 1.1 \\
            Octo-Base \cite{team2024octo}& \ding{55} & 34.7 & 12.5 & 52.8 & 8.3 & 31.9 & 8.3 & 66.7 & 43.1 & 46.5 & 16.0 \\
            Octo-Small \cite{team2024octo}& \ding{55} & 77.8 & 47.2 & 27.8 & 9.7 & 40.3 & 4.2 & 87.5 & 56.9 & 58.4 & 29.5 \\
            OpenVLA \cite{kim2024openvla}& \ding{55} & 4.1 & 0.0 & 33.3 & 0.0 & 12.5 & 0.0 & 8.3 & 4.1 & 14.6 & 1.0 \\
            RoboVLMs \cite{li2024towards}& \ding{51} & 70.8 & 45.8 & 33.3 & 20.8 & 54.2 & 4.2 & 91.7 & 79.2 & 62.5 & 37.5 \\
            CogACT \cite{li2024cogact}& \ding{55} & -- & 71.7 & -- & 50.8 & -- & 15.0 & -- & 67.5 & -- & 51.8 \\
            \makecell[c]{SpatialVLA \cite{qu2025spatialvla}}& \ding{55} & 25.0 & 20.8 & 41.7 & 20.8 & 58.3 & 25.0 & 79.2 & 70.8 & 51.2 & 34.4 \\
            \makecell[c]{SpatialVLA \cite{qu2025spatialvla}}& \ding{51} & 20.8 & 16.7 & 29.2 & 25.0 & 62.5 & 29.2 & \textbf{100.0} & \textbf{100.0} & 53.1 & 42.7 \\
            ThinkAct \cite{thinkact}& \ding{51} & -- & 58.3 & -- & 37.5 & -- & 8.7 & -- & 70.8 & -- & 43.8 \\
            \textbf{FPC-VLA}& \ding{55} & \textbf{79.2} & \textbf{79.2} & \textbf{58.3} & \textbf{58.3} & \textbf{70.8} & \textbf{45.8} & 79.2 & 75.0 & \textbf{71.9} & \textbf{64.6} \\
            \bottomrule
        \end{tabular}
    \end{adjustbox}
    \label{widow}
\end{table*}

\begin{table*}[tb]
\renewcommand\arraystretch{1}
\setlength{\tabcolsep}{3pt}  
\setlength{\abovecaptionskip}{0pt}
\setlength{\belowcaptionskip}{5pt}
\centering
\caption{Comparison results on SIMPLER with Google robot}
\resizebox{1\textwidth}{!}{
\begin{tabular}{ccccccccccccc}
\toprule
\multicolumn{2}{c}{} & \multicolumn{5}{c}{\textbf{Visual Matching}} & \multicolumn{5}{c}{\textbf{Variant Aggregation}} \\
\textbf{Method} & \makecell[c]{\textbf{Fine-tuned}\\\textbf{(in-domain)}} &
\makecell[c]{Pick Coke\\Can (300)} &
\makecell[c]{Move\\Near (240)} &
\makecell[c]{Open/Close\\Drawer (216)} &
\makecell[c]{Open\\Top Drawer\\and Place\\Apple (108)} &
\textbf{Average} &
\makecell[c]{Pick Coke\\Can (825)} &
\makecell[c]{Move\\Near (600)} &
\makecell[c]{Open/Close\\Drawer (378)} &
\makecell[c]{Open\\Top Drawer\\and Place\\Apple (189)} &
\textbf{Average} \\
\midrule
RT-1 \cite{brohan2022rt}& \ding{51} & 85.7 & 44.2 & 73.0 & 6.5 & 52.4 & 89.8 & 50.0 & 32.3 & 2.6 & 43.7 \\
RT-1-X \cite{o2024open}& \ding{55} & 56.7 & 31.7 & 59.7 & 21.3 & 42.4 & 49.0 & 32.3 & 29.4 & 10.1 & 30.2 \\
RT-2-X \cite{o2024open}& \ding{55} & 78.7 & 77.9 & 25.0 & 3.7 & 46.3 & 82.3 & 79.2 & 35.3 & 20.6 & 54.4 \\
Octo-Base \cite{team2024octo}& \ding{55} & 17.0 & 4.2 & 22.7 & 0.0 & 11.0 & 0.6 & 3.1 & 1.1 & 0.0 & 1.2 \\
OpenVLA \cite{kim2024openvla}& \ding{55} & 18.0 & 56.3 & 63.0 & 0.0 & 34.3 & 60.8 & 67.7 & 28.8 & 0.0 & 39.3 \\
RoboVLMs \cite{li2024towards}& \ding{51} & 77.3 & 61.7 & 43.5 & 24.1 & 41.9 & -- & -- & -- &-- & -- \\
CogACT \cite{li2024cogact}& \ding{55} & 91.3 & 80.5 & 71.8 & \textbf{50.9} & 74.8 & 89.6 & 80.8 & 28.3 & 46.6 & 61.3 \\
\makecell[c]{SpatialVLA \cite{qu2025spatialvla}}& \ding{55} & 81.0 & 69.6 & 59.3 & -- & -- & 89.5 & 72.7 & 41.8 & -- & -- \\
    \makecell[c]{SpatialVLA \cite{qu2025spatialvla}}& \ding{51} & 86.0 & 77.9 & 57.4 & -- & -- & 88.0 & 72.7 & 41.8 & -- & -- \\
ThinkAct \cite{thinkact}& \ding{51} & 92.0 & 72.4 & 50.0 & -- & -- & 84.0 & 63.8 & \textbf{47.6} & -- & -- \\
    \textbf{FPC-VLA}& \ding{55} & \textbf{95.3} & \textbf{93.8} & \textbf{76.4} & 46.3 & \textbf{78.0} & \textbf{91.3} & \textbf{86.7} & 30.7 & \textbf{54.5} & \textbf{65.8} \\
    \bottomrule
\end{tabular}}
\label{google}

\end{table*}

\begin{table}[tb]
	\renewcommand\arraystretch{1}
    \setlength{\tabcolsep}{7pt}  
    \setlength{\abovecaptionskip}{0pt}    
    \setlength{\belowcaptionskip}{5pt}
	\centering
	\caption{Comparison results on LIBERO with Franka}
	\resizebox{0.49\textwidth}{!}{
		\begin{tabular}{cccccc}
\toprule
\textbf{Method} & \textbf{Goal} & \textbf{Spatial} & \textbf{Object} & \textbf{Long} & \textbf{Average} \\
\midrule
Diffusion \cite{chi2023diffusion}   & 68.3    & 78.3  & 82.5  & 50.5  & 72.4  \\
Octo-Base \cite{team2024octo}        & 84.6    & 78.9  & 85.7  & 51.1  & 75.1  \\
OpenVLA \cite{kim2024openvla}     & 79.2    & 84.7  & 88.4  & 53.7  & 76.5  \\
TravceVLA \cite{zheng2024tracevla}    & 75.1    & 84.6  & 85.2  & 54.1  & 74.8  \\
SpatialVLA \cite{qu2025spatialvla}  & 78.6    & 88.2  & 89.9  & 55.5  & 78.1  \\
ThinkAct \cite{thinkact} & 87.1 & \textbf{88.3} & 91.4 & 70.9 & 84.4 \\
Cot-VLA \cite{zhao2025cot} & \textbf{87.6} & 87.5 & 91.6 & 69.0 & 81.1 \\
GRAPE \cite{zhang2024grape} & 83.1 & 88.5 & \textbf{92.1} & 57.2 & 80.2 \\
\textbf{FPC-VLA} & 86.2 & 87.0 & 92.0 & \textbf{82.2} & \textbf{86.9} \\
\bottomrule
\end{tabular}}
	\label{libero}
\end{table}

\subsection{Implementation Details}
The VLM model of the supervisor uses Qwen2.5-vl 7B \cite{bai2025qwen2}. During training, we use 16 NVIDIA H100 GPUs. For the VLA model, the batch size is set to 256, the learning rate is $2 \times 10^{-5}$, and 15 action steps are predicted per inference. VLA model is pretrained on the mixture of OXE dataset \cite{o2024open} and LIBERO dataset \cite{liu2023libero}. For the supervisor model, training is conducted with bfloat16 precision, a batch size of 128, a learning rate of $10^{-4}$, LoRA rank of 8, LoRA alpha of 32, and max pixels of 1003520. Synthetic failure dataset is generated from BridgeV2 \cite{walke2023bridgedata} for WidowX, \cite{brohan2022rt} for Google Robot, and \cite{liu2023libero} for Franka. Each robot has a synthetic failure dataset containing 100k entries, while annotated failure dataset generated in MuJoCo and collected from the real robot via teleoperation contain 10k entries. In action fusion module, $\alpha=0.1$, $\lambda=\beta=0.01$. FPC-VLA achieves inference time of 0.176s for non-keyframes and 1.766s for keyframes. As the supervisor is called at most three times during the hundreds of inferences per task, its benefit becomes more evident in long-horizon tasks. For example, in the Open Top Drawer and Place Apple task, the total inference time of FPC-VLA is 40.5 s, which is only slightly higher than the 35.2 s when the supervisor is inactive.

\subsection{Evaluation on Simulation Benchmarks}
To evaluate the FPC-VLA model, we conduct experiments on two simulation platforms—SIMPLER \cite{xuan2024eval} and LIBERO \cite{liu2023libero} —as well as on three types of robotic arms: WidowX, Google Robot, and Franka. We primarily highlight FPC-VLA’s zero-shot learning capability, without domain-specific fine-tuning. The terminology regarding zero-shot and fine-tuning here is consistent with that used in \cite{qu2025spatialvla}. All evaluation tasks follow \cite{zheng2024tracevla,li2024towards,li2024cogact,thinkact}.

\noindent\textbf{Evaluation on SIMPLER \quad}SIMPLER \cite{xuan2024eval} is an open-source suite of simulated environments specifically designed for the comprehensive evaluation of robotic policies. To enhance visual realism and bridge the sim-to-real gap, SIMPLER incorporates two key techniques: Visual Matching and Variant Aggregation. Visual Matching reduces visual disparity between simulation and reality by overlaying real-world background images and fine-tuning object textures so they better align with those found in actual physical environments. Meanwhile, Variant Aggregation introduces multiple randomized variations across visual elements—such as lighting conditions, object appearances, and scene layouts—to increase the diversity and robustness of the training and evaluation settings. In SIMPLER, we test FPC-VLA on two commonly used robotic platforms: WidowX and Google Robot. Visual results can be found in Fig. \ref{viz}.

\begin{figure*}[tb]
  \centering
  \includegraphics[width=1\linewidth, viewport=576 307 1180 487, clip=true]{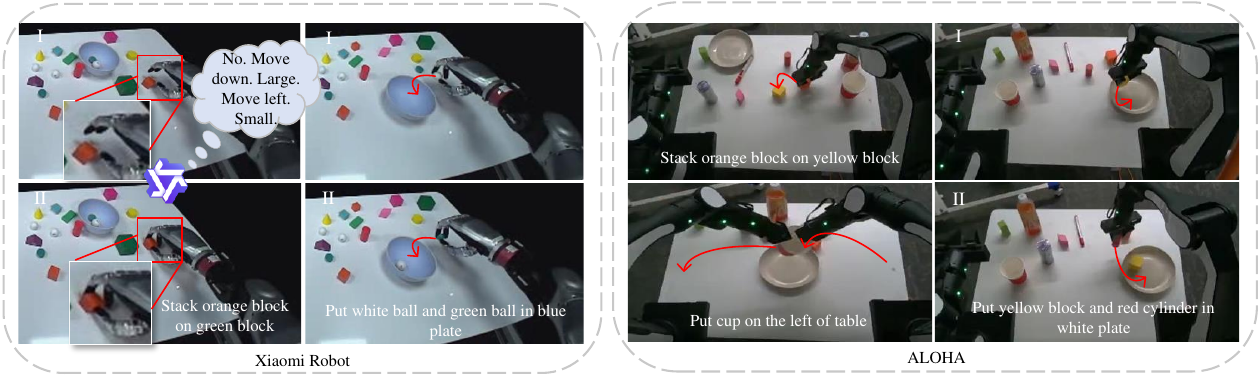}
  \caption{Real world experiments results of FPC-VLA on Xiaomi Robot and ALOHA. Supervisor’s failure correction process is demonstrated taking the task ``Stack orange block on green block" as an example.}
  \label{real}
\end{figure*}

WidowX is evaluated on a suite of manipulation tasks, including grasping individual objects, stacking blocks, and placing items into containers. To assess robustness and generalization, each task is repeated 24 times under diverse initial object poses and environmental configurations. As summarized in Table~\ref{widow}, we compare the performance of several state-of-the-art methods against our proposed FPC-VLA framework, reporting results in terms of grasping success rates and task success rates (\%). Notably, both RT-1-X~\cite{o2024open} and OpenVLA~\cite{kim2024openvla} exhibit poor performance across nearly all tasks, achieving success rates close to 0\%. This suggests significant challenges in transferring policies trained in simulation or on different robot platforms to the WidowX setup, likely due to substantial domain gaps. In contrast, FPC-VLA demonstrates strong zero-shot generalization capabilities—remarkably outperforming even the fine-tuned SpatialVLA~\cite{qu2025spatialvla} without any task-specific adaptation, , demonstrating its effectiveness in both grasping and task completion across varied manipulation scenarios, demonstrating its effectiveness in both grasping and task completion across varied manipulation scenarios.

Google Robot evaluation includes pick-and-place operations and drawer opening/closing tasks. To rigorously assess generalization capabilities, each task is executed repeatedly under diverse conditions—specifically, with varying object poses and the presence of visual distractors. The total number of attempts per task is explicitly reported in parentheses in Table \ref{google}. As shown in Table \ref{google}, in Visual Matching category, FPC-VLA achieves an average score of 78.0\%, outperforming all baselines, despite in zero-shot setting. For Variant Aggregation tasks, which evaluates the ability to generalize across variants, FPC-VLA again ranks highest among all compared methods. It is worth noting that RT-1 \cite{brohan2022rt} is trained solely on Google Fractal dataset \cite{brohan2022rt}, but FPC-VLA significantly outperforms it without any fine-tuning on that dataset. Even when faced with the most complex task (``Open Top Drawer and Place Apple" in Variant Aggregation setting), FPC-VLA still performs remarkably well, surpassing the results of other methods.

\noindent\textbf{Evaluation on LIBERO \quad}
LIBERO \cite{liu2023libero} simulation platform features four distinct task suites: LIBERO-Goal focuses on achieving a variety of task objectives while keeping the set of objects and their spatial arrangements fixed, thereby isolating the challenge of goal-conditioned policy learning. LIBERO-Spatial emphasizes adaptation to varying spatial configurations of the same objects, testing the model’s ability to generalize across different layouts without changes in object identity. LIBERO-Object introduces diversity in the object set itself, requiring robots to interact with a wide range of previously unseen objects, thus evaluating object-level generalization. LIBERO-Long contains long-horizon tasks that involve diverse objects, layouts, and goals.

Following the data modification and augmentation procedures from \cite{kim2024openvla}, we regenerate higher-resolution images (256×256px) , remove no-op actions, rotate third-person images by 180°, filter out failed demonstrations, and use only third-person camera views for fair comparison. Each task suite is evaluated over 500 trials. As shown in Table \ref{libero}, FPC-VLA achieves the highest average performance of 86.9\%, significantly outperforming all other baselines despite not using in-domain data for fine-tuning. Notably, on the most challenging long-horizon tasks, it scores 82.2\%, well above the second-best 70.9\% by ThinkAct \cite{thinkact}, indicating that failure prediction and correction can help avoid deviation in long-term tasks.

\subsection{Real-World Experiments}
To assess FPC-VLA’s real-world applicability, as shown in Fig. \ref{real}, we evaluate five challenging tasks on the Xiaomi Robot and ALOHA.

\begin{figure}[tb]
  \centering
  \includegraphics[width=1\linewidth, viewport=59 311 481 512, clip=true]{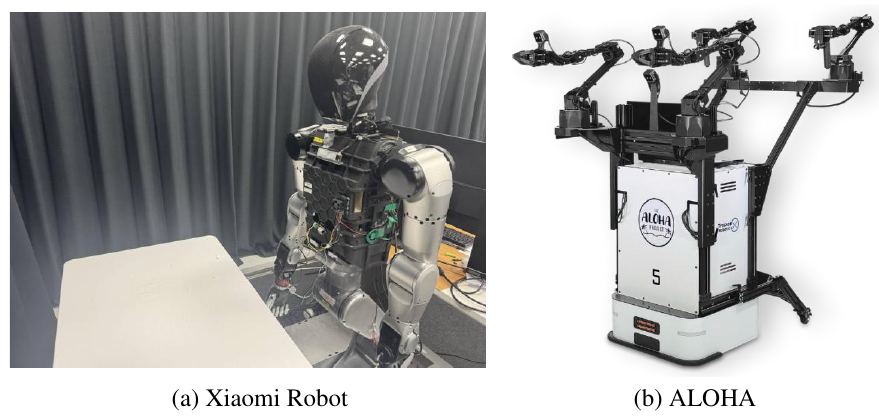}
  \caption{Xiaomi Robot and ALOHA for real world experiments.}
  \label{robot}
\end{figure}

\begin{table*}[tb]
    \renewcommand\arraystretch{1.2}
    \setlength{\abovecaptionskip}{0pt}    
    \setlength{\belowcaptionskip}{5pt}
	\centering
	\caption{Comparison results in real world experiments}
	\resizebox{1\textwidth}{!}{
		\begin{tabular}{ccccc|cccccc|cc}
\toprule
\textbf{Robot} & \multicolumn{4}{c}{\textbf{Xiaomi Robot}} & \multicolumn{6}{c}{\textbf{ALOHA Robot}} & \multirow{3}{*}{\textbf{Avg. Grasp}} & 
\multirow{3}{*}{\textbf{Avg. Task}}  \\
\multirow{2}{*}{\textbf{Method}} &
\multicolumn{2}{c}{\makecell{\textbf{Stack orange block}\\ \textbf{ on green block}}} & 
\multicolumn{2}{c}{\makecell{\textbf{Put white ball}\\ \textbf{and green ball} \\\textbf{in blue plate}}} & 
\multicolumn{2}{c}{\makecell{\textbf{Stack orange block}\\\textbf{on yellow block}}} & 
\multicolumn{2}{c}{\makecell{\textbf{Put cup on}\\\textbf{the left of table}}} &
\multicolumn{2}{c}{\makecell{\textbf{Put yellow block}\\ \textbf{and red cylinder}\\ \textbf{in white plate}}}&
\\
& Grasp & Task 
 & Grasp & Task 
 & Grasp & Task 
 & Grasp & Task  & Grasp & Task 
 & & \\
\midrule
ACT \cite{zhao2023learningfinegrainedbimanualmanipulation}& 62 & 34 & 58 & 52 & 64 & 30 & 72 & 58 & 52 & 46 & 61.6 & 44.0\\
Diffusion \cite{chi2023diffusion}& 44 & 12 & 22 & 18 & 14 & 8 & 36 & 26 & 38 & 32 & 30.8 & 19.2 \\
OpenVLA \cite{kim2024openvla}& 42 & 22 & 38 & 36 & 54 & 28 & 56 & 42 & 28 & 24 & 43.6 & 30.4\\
$\pi_0$ \cite{black2024pi0visionlanguageactionflowmodel}& 58 & 42 & 60 & 54 & 72 & 32 & 44 & 36 & 40 & 34 & 54.8 & 39.6\\
$\pi_0$-Fast \cite{pertsch2025fastefficientactiontokenization}& 38 & 12 & 44 & 42 & 54 & 40 & 60 & 48 & 46 & 38 & 48.4 & 36.0\\
CogACT \cite{li2024cogact} & 82 & 70 &86 & 82 & 88 & 78 & 90 & 82 & 84 & 80 & 86.0 & 78.4\\
\textbf{FPC-VLA}& \textbf{96} & \textbf{84} & \textbf{88} & \textbf{88} & \textbf{94} & \textbf{82} & \textbf{96} & \textbf{90} & \textbf{90} & \textbf{86} & \textbf{92.8} & \textbf{86.0} \\
\bottomrule
\end{tabular}}
	\label{realworld}
\end{table*}


\noindent\textbf{Xiaomi Robot Setup \quad}Fig.\ref{robot}(a) illustrates our 7-DoF Xiaomi Robot setup. The head is equipped with an Intel RealSense D435 RGB-D camera, which captures close-range task observations by streaming 640×480 RGB-D frames at 30 Hz. This configuration supports real-time workspace perception, enabling capabilities such as object detection and pose estimation. All policy outputs are provided as Cartesian coordinate increments expressed in the robot’s base frame, facilitating seamless integration with the arm’s control system through a ROS-based interface. Data collection occurs at 10 Hz via teleoperation, ensuring compatibility with low-latency applications.

\noindent\textbf{ALOHA Robot Setup \quad}Our ALOHA hardware platform (Fig. \ref{robot}(b)) is built upon the Tsinghua Airbot system. It features two lightweight 6-degree-of-freedom (6-DoF) manipulators mounted on a compact mobile base. Each arm is fitted with a 1-DoF parallel gripper and a wrist-mounted RGB-D camera to support close-range manipulation perception. Additionally, an Intel RealSense D435 RGB-D camera is attached to an adjustable overhead mast, capturing RGB-D frames at a resolution of 640×480 and a rate of 30 Hz.  Whenever the pose of the overhead camera changes, we perform hand-eye calibration using an AprilTag calibration board to estimate the rigid transformation between the robot base frame and the camera frame. During preprocessing, all teleoperated trajectories are transformed into the overhead camera’s coordinate frame. Conversely, actions generated by the learned policy are mapped back to the robot base frame using the calibrated extrinsic parameters.  The model outputs Cartesian incremental commands, which are integrated over time to produce target end-effector poses. These poses are then converted into joint-space commands via a damped least-squares inverse kinematics solver, with explicit enforcement of joint velocity and acceleration limits.

\noindent\textbf{Real World Data Preparation \quad}For the Xiaomi Robot, we collect data by controlling the robotic arm using a SpaceMouse. For ALOHA, we use its built-in dual-arm teleoperation system to collect trajectories containing joint positions, end-effector poses, and RGB-D images. The raw data undergoes several preprocessing steps: first, we apply a median filter with a kernel size of 5 and remove outlier frames exceeding a threshold of three standard deviations to eliminate noise; second, we temporally synchronize the multimodal data streams and downsample the ALOHA data from 30 Hz to 10 Hz to match the temporal frequency of the Xiaomi robot data, ensuring temporal alignment across datasets; finally, we convert the processed data into the RLDS format, structuring it to align with the OXE dataset schema, which includes standardized fields for observations, actions, and rewards.

\noindent\textbf{Analysis of Comparison Results \quad}Table \ref{realworld} presents a comprehensive comparison of various robotic manipulation methods across multiple real-world tasks performed on two distinct robotic platforms: the Xiaomi Robot and the ALOHA Robot. The evaluation metrics include grasp success rate (Grasp) and task completion success rate (Success). Inference uses FP16 on a single NVIDIA RTX 3090, with 50 trials per task. Our proposed method, FPC-VLA, consistently outperforms all baseline approaches across both robots and all individual tasks. While $\pi_0$ [45] and OpenVLA [15] show moderate performance, methods like Diffusion [40] and $\pi_0$-Fast [46] struggle with low success rates, especially on the ALOHA platform. ACT [44] performs reasonably well but is still substantially outperformed by FPC-VLA in both grasp precision and task completion reliability. FPC-VLA achieves the highest average grasp success rate (92.8\%) and task success rate (86.0\%), significantly surpassing the second-best method, CogACT [19], which attains 86.0\% and 78.4\%, respectively. As shown in Fig. \ref{real}, thanks to the VLM-based Supervisor, FPC-VLA demonstrates strong failure prediction and correction capabilities, effectively reducing instances of grasping or placement failures. Notably, we include a dual-arm handover task (Put cup on the left of table), which demonstrates FPC-VLA’s versatility across multi-configuration robots.

\begin{figure}[tb]
  \centering
  \includegraphics[width=1\linewidth]{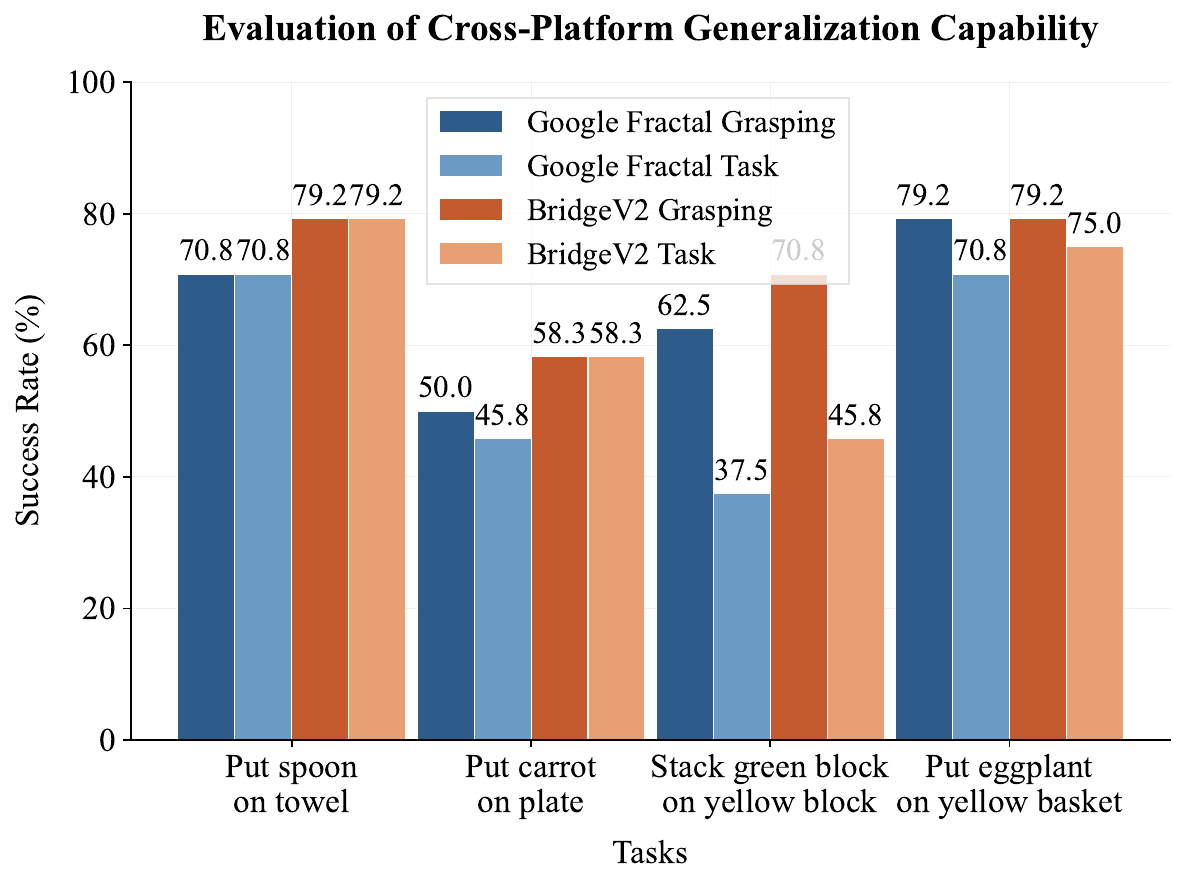}
  \caption{Evaluation of Supervisor cross-platform capability. Success rates (\%) are reported for models trained on Google Fractal and BridgeV2 datasets, and evaluated on WidowX robot.}
  \label{cross}
\end{figure}

\subsection{Evaluation of Cross-Platform Generalization Capability}
To demonstrate the cross-platform generalization capability of our supervisor, we conduct experiments on the WidowX robot in SIMPLER using a supervisor fine-tuned exclusively on data from \cite{brohan2022rt} (Google Robot). This setting evaluates whether the failure prediction and correction policies learned on one robotic platform can effectively transfer to another with different morphology, actuation, and visual appearance—without any additional fine-tuning or adaptation. The results, presented in Fig. \ref{cross}, show that our supervisor achieves competitive success rates across all evaluated tasks on the WidowX robot, despite being trained solely on Google Robot data. The remarkable zero-shot performance arises from several synergistic design principles in FPC-VLA: First, the supervisor reasons over high-level, task-grounded natural language queries instead of low-level robot states, decoupling failure analysis from embodiment details. Second, corrections use a standardized, egocentric vocabulary of relative directions and magnitudes, enabling platform-agnostic transferability compared to absolute action deltas. Thirdly, the failure-correction dataset encodes only task-relevant spatial changes into language, capturing universal manipulation heuristics that generalize across robot embodiments.

\begin{figure}[tb]
  \centering
  \includegraphics[width=1\linewidth]{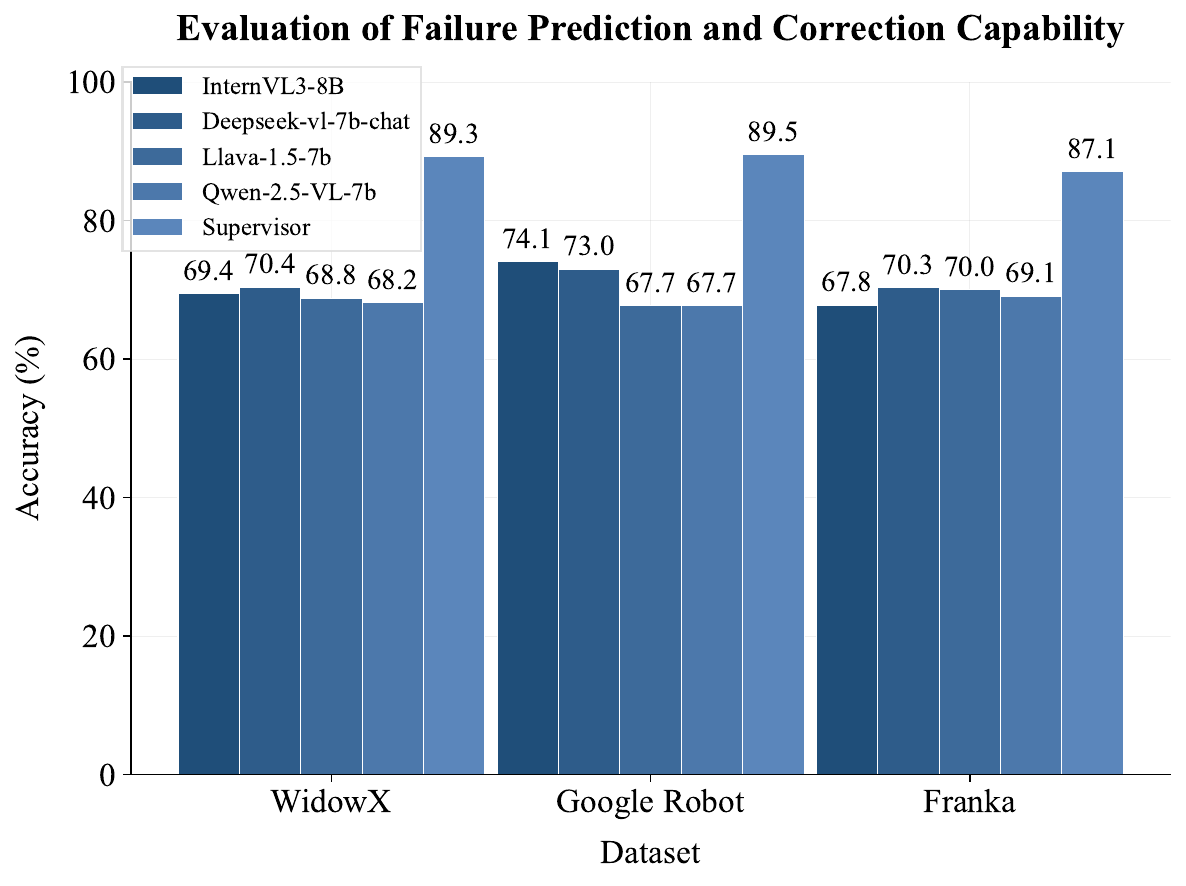}
  \caption{Evaluation of failure prediction and correction accuracy between our supervisor and other advanced VLMs.}
  \label{acc}
\end{figure}

\subsection{Evaluation of Failure Prediction and Correction}
To evaluate the accuracy of our supervisor’s failure prediction and correction capability, we conduct a comparative study against several state-of-the-art vision–language models (VLMs), including DeepSeek-VL-7B \cite{zhu2024deepseek}, InternVL3-8B \cite{zhu2025internvl3}, LLaVA-1.5-7B \cite{liu2024improved}, and the base Qwen2.5-VL-7B \cite{bai2025qwen2}. Following standard practice for evaluating task-specific adaptation, our supervisor is fine-tuned on the automatically generated failure correction dataset described in Section \ref{dataset}, which provides structured vision–language–action supervision tailored to robotic manipulation failures. In contrast, the baseline VLMs are evaluated using their publicly released, general-purpose pretrained weights without any task-specific fine-tuning or prompt engineering beyond what is necessary to align with our input format.

\begin{table*}[tb]
	\renewcommand\arraystretch{1}
    \setlength{\tabcolsep}{4pt}  
    \setlength{\abovecaptionskip}{0pt}    
    \setlength{\belowcaptionskip}{5pt}
	\centering
	\caption{Results of ablation study}
	\resizebox{1\textwidth}{!}{
		\begin{tabular}{cccccccccccccc}
\toprule
\multicolumn{2}{c}{\textbf{Robot}} & \multicolumn{4}{c}{\textbf{WidowX}} & \multicolumn{2}{c}{\textbf{Xiaomi robot}} & \multicolumn{2}{c}{\textbf{ALOHA}} & \multirow{3}{*}{\textbf{Avg. Grasp}} & 
\multirow{3}{*}{\textbf{Avg. Task}} \\
\multirow{2}{*}{\textbf{Module}} & \multirow{2}{*}{\textbf{Setting}} & 
\multicolumn{2}{c}{\makecell{\textbf{Put spoon}\\\textbf{on towel}}} & 
\multicolumn{2}{c}{\makecell{\textbf{Put eggplant}\\\textbf{on yellow basket}}} & 
\multicolumn{2}{c}{\makecell{\textbf{Stack orange block}\\\textbf{on green block}}} & 
\multicolumn{2}{c}{\makecell{\textbf{Put cup on}\\\textbf{the left of table}}} & 
 \\
& & Grasp & Task 
 & Grasp & Task 
 & Grasp & Task 
 & Grasp & Task \\
 
\midrule
\multirow{3}{*}{Supervisor} & w/o Supervisor & 70.8 & 70.8 & 79.2 & 70.8 & 84 & 72 & 90 & 84 & 81.0 & 74.4 \\
& w/o correction & 75.0 & 70.8 & \textbf{83.3} & 70.8 & 84 & 74 & 92 & 86 & 83.6 & 75.4\\
& \makecell{w/o example samples} & 75.0 & 70.8 & \textbf{83.3} & 75.0 & 94 & 82 & 92 & 88 & 86.1 & 79.0 \\
\midrule
\multirow{3}{*}{\makecell{Action Fusion\\Module}} & \makecell{use latest prediction} & 29.2 & 12.5 & 41.7 & 33.3 & 52 & 40 & 72 & 66 & 48.7 & 38.0\\
& directly average & 79.2 & 70.8 & \textbf{83.3} & 66.7 & 84 & 76 & 86 & 82 & 83.1 & 73.9\\
& w/o time decay & 79.2 & 79.2 & 79.2 & 66.7 & 92 & 82 & 94 & 90 & 86.1 & 79.5\\
\midrule
\multicolumn{2}{c}{motion disturbance + w/o Supervisor} & 62.5 & 54.2 & 37.5 & 25.0 & 72 & 60 & 72 & 64 & 61.0 & 50.8\\
\multicolumn{2}{c}{motion disturbance + FPC-VLA} & 75.0 & 62.5 & 62.5 & 50.0 & 80 & 74 & 82 & 78 & 74.9 & 66.1\\
\midrule
\multicolumn{2}{c}{\textbf{FPC-VLA}} & \textbf{79.2} & \textbf{79.2} & 79.2 & \textbf{75.0} & \textbf{96} & \textbf{84} & \textbf{96} & \textbf{90} & \textbf{87.6} & \textbf{82.1}\\
\bottomrule
\end{tabular}}
	\label{ablation}
\end{table*}

As shown in Fig. \ref{acc}, our fine-tuned supervisor consistently outperforms all baselines on WidowX, Google Robot and Franka. This performance gap highlights the effectiveness of both our data generation pipeline and the alignment between the supervision signal and the target task. Importantly, even when compared to its own base architecture Qwen2.5-VL-7B, our approach achieves a significant improvement solely through fine-tuning on synthetically generated, robotically grounded QA pairs. Our results suggest that while general-purpose VLMs possess strong visual and linguistic capabilities, they benefit substantially from targeted fine-tuning on semantically structured, robot-relevant data—even if synthesized—to achieve reliable performance in safety-critical decision-making contexts.


\subsection{Ablation Study}
Table \ref{ablation} presents the ablation study on FPC-VLA, focusing on the Supervisor and Action Fusion Module. ``w/o" indicates ``without". Results are measured on WidowX, Xiaomi robot and ALOHA tasks. Removing the Supervisor greatly degrades performance, as it corrects deviations from the expert trajectory. Removing corrections during data generation and inference also lowers success, indicating that VLM’s correction suggestions aid action optimization. Providing in-context examples further improves success by clarifying task requirements and output format. For the Action Fusion Module, similarity-guided fusion outperforms both no fusion and simple averaging by better leveraging multi-modal actions, while the time-decay strategy boosts success by emphasizing recent predictions.

To further evaluate the Supervisor’s effect on robustness, we perform an action perturbation experiment with and without the Supervisor. The end-effector pose is randomly disturbed within the range of [0.01, 0.1]. Results show that although  perturbations reduce success rates in both settings, the drop is 31.3\% without the Supervisor but only 16\% with FPC-VLA.

\section{Conclusion}
In this paper, we propose FPC-VLA, a VLA Framework with a supervisor for failure prediction and correction. By integrating a VLM-based supervisor and a dual-stream action fusion module, FPC-VLA not only refines robot actions before execution but also proactively mitigates potential failures. Extensive experiments across simulated environments and real-world robotic platforms demonstrate that FPC-VLA achieves superior performance over existing methods in zero-shot and fine-tuned settings, highlighting its robustness, generalization ability, and potential for safe and reliable deployment in open-ended robotic applications.

\section{Limitations and Future Work}
While FPC-VLA demonstrates strong performance in failure prediction and correction, it has some limitations. First, the supervisor is triggered exclusively by gripper state changes, making it unable to address post-grasp failures—such as objects slipping or dropping after successful pickup—which are common in real-world manipulation. Mitigating such issues typically requires force/torque sensing and compliant control strategies, yet current VLA frameworks, including ours, rely solely on visual inputs and lack the necessary modalities to perceive or react to contact forces. Integrating force sensors into the VLA pipeline could enable more robust handling of dynamic interactions. Second, although our experiments include long-horizon tasks (e.g., drawer opening and stove operation), their complexity and duration remain limited compared to the demands of real-world domestic service scenarios, which often involve hours of continuous, multi-step reasoning under uncertainty. Nevertheless, as VLA architectures continue to evolve—with improvements in memory, planning, and multimodal grounding—we believe this paradigm holds significant promise for scaling to truly complex, everyday robotic applications.








\bibliographystyle{elsarticle-num}

\bibliography{ref} 
\end{document}